\documentclass[runningheads]{llncs}
\usepackage{graphicx}
\usepackage[utf8x]{inputenc}
\usepackage{array}
\usepackage{comment}
\begin{document}
\title{Datacentric analysis to reduce pedestrians accidents: A case study in Colombia}
%Danger factor in Patterns Behavior of pedestrians crossing streets
\titlerunning{Pedestrian Modeling}
\author{Michael Puentes\inst{1}\orcidID{0000-0002-1802-839X}\and
Diana Novoa\inst{2} \and
John M. Delgado Nivia\inst{3} \and
Carlos J. Barrios Hernández\inst{1} \orcidID{0000-0002-3227-8651} \and
Oscar Carrillo\inst{4} \orcidID{0000-0001-5081-1774} \and 
Frédéric Le Mouël\inst{5}\orcidID{0000-0002-7323-4057} }
\authorrunning{UIS - UTS - CPE - INSA}
\institute{Universidad Industrial de Santander
\email{michael.puentes@correo.uis.edu.co,cbarrios@uis.edu.co}
\and
Universidad nacional de la plata
\email{diananovoa4@gmail.com}
\and
Unidades Tecnológicas de Santander
\email{territoriointeligente@uts.edu.co} 
\and
Univ Lyon, CPE, INSA Lyon, Inria, CITI, EA3720, F-69621 Villeurbanne, France
\email{oscar.carrillo@cpe.fr}
\and
Univ Lyon, INSA Lyon, Inria, CITI, EA3720, F-69621 Villeurbanne, France \email{frederic.le-mouel@insa-lyon.fr}}
\maketitle              % typeset the header of the contribution
\begin{abstract}
Since 2012, in a case-study in Bucaramanga-Colombia, 179 pedestrians died in car accidents, and another 2873 pedestrians were injured. Each day, at least one passerby is involved in a tragedy. Knowing the causes to decrease accidents is crucial, and using system-dynamics to reproduce the collisions' events is critical to prevent further accidents. This work implements simulations to save lives by reducing the city's accidental rate and suggesting new safety policies to implement. Simulation's inputs are video recordings in some areas of the city. Deep Learning analysis of the images results in the segmentation of the different objects in the scene, and an interaction model identifies the primary reasons which prevail in the pedestrians or vehicles' behaviours. The first and most efficient safety policy to implement - validated by our simulations - would be to build speed bumps in specific places before the crossings reducing the accident rate by 80\%.

\keywords{Data-centric  \and traffic violation \and Dynamic System}
\end{abstract}
\section{Introduction}
%\subsection{A Subsection Sample}
This work aims to reduce accidents in a city by knowing pedestrian behaviour in a city's urban area. 

Yang et al. \cite{Yang2006q} contributed notably to the field by separating two kinds of pedestrians - those obeying the law and those having opportunistic behaviours. Under this assumption, Yang made a questionnaire evaluating Chinese citizens' behaviour related to the pedestrian cross path. In his survey, some variables are essential in the model construction for micro-simulation, like age and gender. This characteristic is relevant according to the perception of the pedestrians in the case study in Colombia \cite{TorG:CiuRSCPDAVCP,PuenM:TowSCICMFGUCTCS}. Indeed, pedestrians clearly show in our dataset an obedience issue to the traffic authorities. Another related work by Chen Chai et al. \cite{Chen2016} evaluates pedestrians' behaviour also by gender and age, but adds extra information from fuzzy logic-based observations about if the subject is a child. This work question the indicators that influence pedestrians' behaviour. Aaron et al. \cite{Aaron2019} compare real situations where different variables related to the environment were evaluated, knowing the reality will always be different. This work defines the variables of a micro-simulation which have to be part of the causal model to refine the modelled reality. Camara et al. \cite{Camara2018} implement a decision tree to determine the pedestrians' vehicle interaction. This implementation, looking for the design of new policies for pedestrians in Bucaramanga - Colombia, seeks for less critical accidents in 2019 over 200 pedestrians involved in an accident\footnote{https://www.datos.gov.co/Transporte/Accidentes-de-Transito-en-Bucaramanga-ocurridos-de/7cci-nqqb}. They identify the need to find the correct variables that can reduce pedestrians accidents. For instance, Holland et al. \cite{Holland2010} highlight gender as an essential factor in the behavior of a pedestrian to decide a crossing. Other works consider pedestrian and vehicles flows \cite{oskarbski2018,Campisi2018}, even if the use of specific PTV-VISSIM and VISWALK software modules\footnote{http://vision-traffic.ptvgroup.com/es/productos/ptv-vissim/} can recognize pedestrian events and drivers’ behaviour as individuals.

These different works well-illustrate the importance of determining the exact criteria influencing pedestrians and drivers to take an action at a crossing intersection. In section \ref{PreviuosWork}, we go through several causes found in related works. Then, in section \ref{Methodology}, we describe the methodology used in our work to build the dataset and the simulation. In section \ref{Model}, we detail our model and simulation results. We finally conclude and give future works.

\section{Related work} \label{PreviuosWork}
\begin{figure}[ht]
\centering
\includegraphics[width=0.6\textwidth]{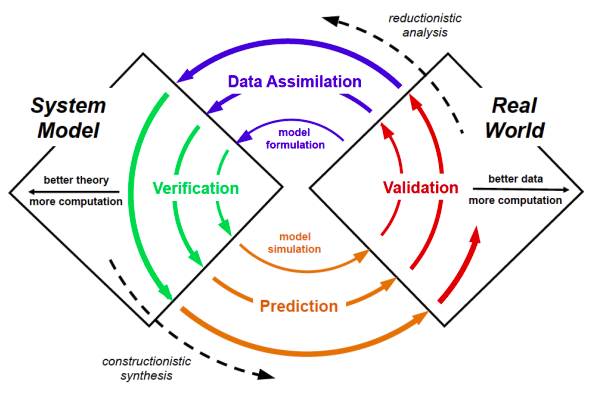}
\caption{Inference Spiral of System Science (Jordan 2015) }
\label{fig:i_jordan}
\end{figure}

%%%%%%%%%%%%%%%%%%%%%%%%%%%%%%%%%%%%%%%%%%%%%%%%%%%%%%%%
Several research works try to identify the variables intervening in pedestrians' jaywalking producing accidents in a city  - either from pedestrian behaviour or external causes. To find these causes, we establish different scenarios with real-life information and people perception. To validate these variables in our micro-simulation model, we refine them by comparing their value with video-recordings in a spiral process of refined simulation by Jordan (see Fig \ref{fig:i_jordan}) \cite{PuenM:PedBMSRTDI}. This validation is repeatedly processed in an endless task - as citizens behaviours are not trivial and only a near prediction is possible. It allows identifying new potential variables to be in the model of the citizens' perception. Furthermore, to compare the results with existing models \cite{Cantillo2015}.

To minimize the accidents that involved citizens and vehicles, it is necessary to find the reasons that affect pedestrians' opportunistic decisions when the traffic light allows the pass of the vehicles or scenarios where do not exist signals. Pedestrians must have priority in the way, and the vehicle has to stop in that case. The empirical analysis of Sanghamitra et al. \cite{das2005} lists the crossing decisions of passersby in a cross-side of the street based on the time gap until the next car. Even if all the variables could not be detected by this research tool, it is possible to recognize similar walkers' behaviours. Another well-known variable taking place in pedestrian behaviour is the "social force". It occurs when the pedestrians are guided by another citizen, without knowing if the principal citizen decision is correct, but at least having a partial vision of the path. Different passersby models can be found in the literature: magnetic force, social force, and Benefit-Cost Cellular. Teknomo does a review of microscopic simulations of pedestrians, detailing every pedestrian as an individual \cite{teknomo2016}. In his work, different variables are necessary to the mathematical modelling, but different causes could be part of a pedestrian accident. In a simulation, several causes can be considered in the model:

\begin{itemize} 
\item Time gap between car and pedestrian \cite{Yang2006q,das2005}
\item Social force \cite{teknomo2016,Helbing1995}
\item Environment (weather, pollution, noise) \cite{Kumar2011}
\item Vehicle factors \cite{Kumar2011}
\item Human factors (driver skills, fatigue, alcohol, drugs, too quick glance) \cite{Kumar2011,Cookson2011}
\item Road Conditions (corner, visibility, straight, wet, dry)     \cite{Kouabenan2005}
\end{itemize}

These causes allow identifying different scenarios where a pedestrian have a specific behaviour. Pau et al. \cite{Pau2018} select scenarios at different times of the day: with many pedestrians in the streets (peak hours) or when almost no pedestrians walk through the street. Rasouli et al. \cite{Rasouli2018} determine the people who cross-traffic line and the street and who made a signal with the hands, showing a petition of the stop to the driver. This article's methodology looks at different environments: night, day, rain, snow - in the same place when crowded or not. This work is not related to implementing objects in the video sequence, but when the behaviour of pedestrian changes according to luminosity, it is a parameter we need to consider. Kouabenan et al. \cite{Kouabenan2005} analyze 55 reports of pedestrians' accidents, randomly selected from a police report on the IvoryCoast. In their article, they analyze the characteristics and circumstances of the accidents. Those previous works conclude their research by suggesting alternative solutions, simplifying or new public policies, campaigns, or improvements in a particular spot of the city. Proposed actions are:

\begin{itemize} 
\item Accident prevention campaigns \cite{Kouabenan2005,Mendez2014}
\item Road safety policies \cite{Kumar2011,Ulfarsson2010} %20 percent of reduction according to Kumar2011
\item Improved lighting conditions \cite{Ulfarsson2010,Zhang2014}
\item Vehicle conditions campaigns \cite{Ulfarsson2010}
\end{itemize}

%%%%%%%%%%%%%%%%%%%%%%%%%%%%%%%%%%%%%%%%%%%%%%%%%%%%%%%
\section{Methodology} \label{Methodology}

Our research work aims to create a micro-simulation model that considers factors that interfere in the citizen's behaviour (pedestrian or driver). Later on, the simulation is refined with real-life video data from cameras installed in different spots in five cities in Colombia. This work is then validated with a case study on the city of Bucaramanga by using the paradigm of system dynamics; the model evolves by applying the following four steps:

\begin{enumerate} 
\item Identify and analyse the factors influencing on the event 
\item Modify the parameters of the simulation model
\item Execute the simulation with the modified model % Evaluate the equations created from here
\item Observe and differentiate between the real-life behaviour on the observed event and the simulation outcomes
\end{enumerate}

\begin{figure}[ht]
\centering
\includegraphics[width=0.7\textwidth]{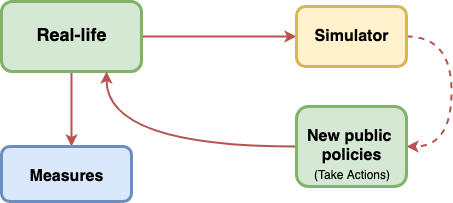}
\caption{General methodology implemented }
\label{fig:i_general_method}
\end{figure}

The process previously described is a cycle that improves the model as soon as we add more variables according to the phenomenon's observations (see Fig~\ref{fig:i_general_method}). If it is possible to find the factors that significantly impact the pedestrians' accidents in the city, these causes can become the targeted part of the new public policies to reduce future tragedies.

To validate our model and apply our methodology, we collaborate with the Santander department's government in Colombia. 
One of the projects - funded by the government - aims to analyse the citizens' behaviour and improve civics in the city. To this end, 900 cameras were deployed in five cities of the department.
Different projects are proposed to use these cameras:  mobility, public spaces and harmony. 

\begin{figure}[ht]
\centering
\includegraphics[width=0.8\textwidth]{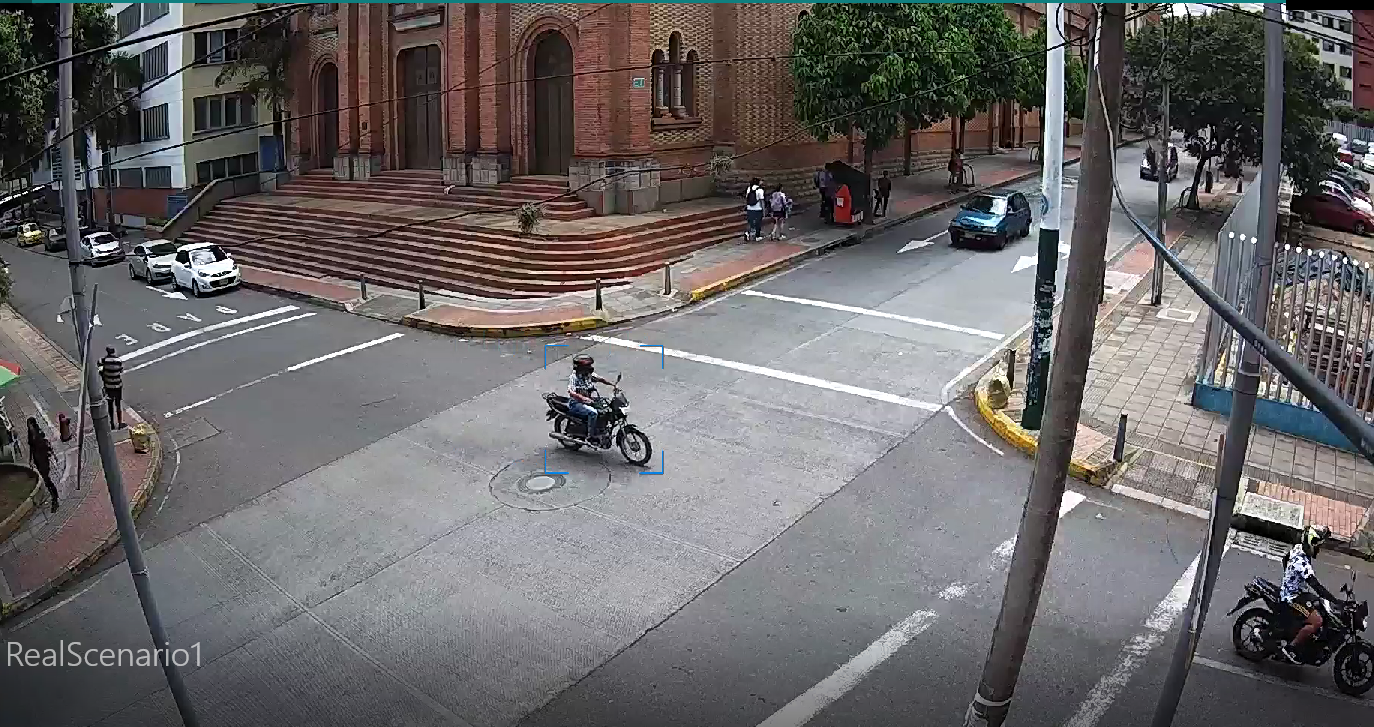}
\caption{Camera selected in Bucaramanga-Colombia }
\label{fig:i_moto_atravezada}
\end{figure}

This research focuses on pedestrians to further a better mobility and reduce the rate of accidents in the city. Thus we select, from the deployed cameras, the spots where more accidents in particular conditions occur. An image of one of the top five selected spots where collisions happen very often is shown in Figure \ref{fig:i_moto_atravezada}. 
This particular is in a relevant neighbourhood collecting different variables according to the related works (see Section \ref{PreviuosWork}): no traffic lights, the second area in the city with more accidents, presenting architecture heterogeneity: one church, two universities, one park, and numerous residential areas nearby.

The initial simulation model starts from the previous research carried out on the evaluated phenomenon, bearing in mind that all the models can vary because we created the model for the particular case of Bucaramanga. The real-life factors can be analysed and measured thanks to the cameras of the city. To measure these factors, we analyse hours of video on the Briefcam\footnote{https://www.briefcam.com/} software. This software has tools to measure causes and find objects via deep learning on video-recordings and accelerate the process of identifying pedestrians. Then, with the video-recordings and the software, it is possible to get the data detailed in Table~\ref{table:ListOfVariables}. This report is necessary to have extra information from particular behaviours from the video and the perception of the people who live in Bucaramanga.

\setlength{\tabcolsep}{18pt}
\renewcommand{\arraystretch}{1.5}

%%%%%% TABLE %%%%%%%
\begin{table}[h!]
\centering
\begin{tabular}{ | m{12em} | m{5em}| m{10em} | } 
 \hline
 \textbf{Variable} &  \textbf{Type} & \textbf{Observation}\\ 
 \hline\hline
 Date of Video  & Date  &  \\
 Range hour &   e.g: 21:00 - 22:00 hrs & \\
 Pedestrians who cross the street  &   numeric  & categorized by gender \\
 Not safe events  &   numeric & those are events which involves an accident \\
 Pedestrians against the law  &   numeric & \\
 Pedestrians who cross by the zebra  &   numeric & \\
 Pedestrians velocity average  &   numeric & \\
 \hline
\end{tabular}
\caption{Variables describing the pedestrians' behaviours}
\label{table:ListOfVariables}
\end{table}

\section{Model and simulation results} \label{Model}
%%%%%%%%%%%%%%%%%%%%%%%%%%%%%%%%%%%%%%%%%%%%%%%%%%%%%%%

\begin{figure}[ht]
\centering
\includegraphics[width=0.8\textwidth]{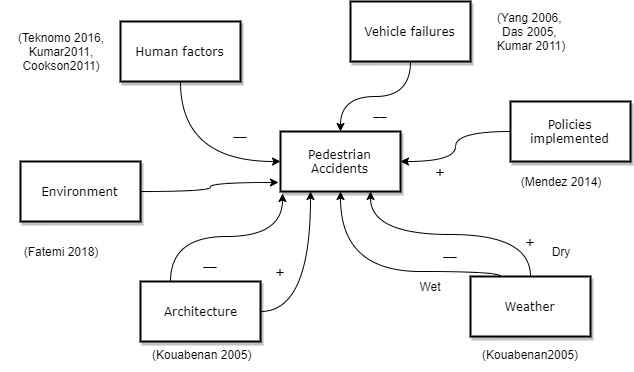}
\caption{Causal diagram for pedestrian accidents}
\label{fig:i_causal_diagram}
\end{figure}

According to the related works reviewed in Section \ref{PreviuosWork}, some causes directly entailing pedestrians’ accidents are observable. Otherwise, for non-trivial causes, we estimate the necessity of micro-simulation \cite{AlvarezPomar2016}: seeing the pedestrian as an individual, identifying the general causes seen in the observed simulation, and measurable in the cameras installed for this project (see Fig. \ref{fig:i_causal_diagram}). In previous researches, many simulations were performed \cite{Cantillo2015,Kumar2011,Rasouli2018,Mendez2014,Ulfarsson2010}, and the proposed solutions produced new policies, which only have a sinusoidal behaviour, according to Mendez \cite{Mendez2014}. Therefore, it is necessary to implement a micro-simulation to identify and analyse the pedestrians’ causes of accident as an individual \cite{Yang2006q,teknomo2016} and determine the particular factors that will reduce the accident rate. %, or as an example unnecessary a walker, but between vehicles as the image \ref{fig:i_moto_atravezada}.

%IMAGE

With micro-simulation, factors can directly be compared to measurable variables thanks to video recordings from the initial causes evaluated. In particular, this work assesses human factors in the micro-simulation for pedestrians as individuals. 
The numeric data is not enough. Hence we use the citizens' perception to have extra information not visible in the videos.

In some accidents, pedestrians endanger their own lives, and thus they interfere with traffic. In this simulation, we aim to look for a safer pedestrian crossing. Therefore, we use the Viswalk simulation tool\footnote{https://www.ptvgroup.com} to find the causes of accidents in a micro-simulation of walkers in a specific sector of Bucaramanga (Colombia), using the same method represented in Fig \ref{fig:i_general_method}. 

\begin{figure}[ht]
\centering
\includegraphics[width=0.8\textwidth]{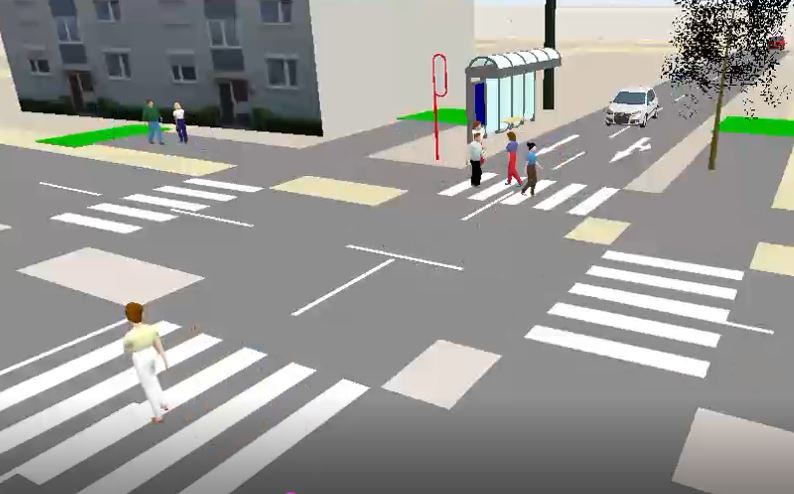}
\caption{Viswalk micro-simulation in San Francisco neighborhood}
\label{fig:i_cruce_canaveral}
\end{figure}

Firstly, we analyse the priority, which in Colombia is for vehicles instead of pedestrians.
To know the causes that a jaywalker could have, we interviewed students and other passers-by about how good pedestrians they were. The results are shown in Tables \ref{table:Answers} and \ref{table:AnswersProbabilities}. This information was used to feed the micro-simulation, for example, and one of the more exciting information, \emph{69\% of people think that the pedestrian is disrespected}, even in a zebra crossing, the car or motorbike has priority. This is important information, when the simulation has this rule of priority, the accidents in the micro-simulation appear (see Figure \ref{fig:i_cruce_canaveral}).

%%%%%% TABLE %%%%%%%
\begin{table}[h!]
\centering
\begin{tabular}{ | m{12em}  m{1.5em} m{1.5em} m{2em} m{1.5em} m{1.5em}| } 
 \hline
 \textbf{Question} &  \textbf{always} & \textbf{often} & \textbf{sometimes} & \textbf{rarely} & \textbf{never}\\ 
 \hline\hline
 Do you walk on the zebra crossing when you cross the road? & 42.9\%  & 47.6\% & 9.5\% & 0.0\% & 0.0\%  \\
Do you look at the state of traffic lights when you cross the road? & 95.2\%  & 4.8\% & 0.0\% & 0.0\% & 0.0\%  \\

 \hline
\end{tabular}
\caption{Pedestrians' behavior at intersections}
\label{table:Answers}
\end{table}

%%%%%% TABLE %%%%%%%
\begin{table}[h!]
\centering
\begin{tabular}{ | m{12em}  m{1.5em} m{1.5em} m{2em} m{2em} m{1.5em}| } 
 \hline
 \textbf{Question} &  \textbf{always} & \textbf{often} & \textbf{sometimes} & \textbf{rarely} & \textbf{never}\\ 
 \hline\hline
 In general conditions & 0.0\%  & 9.5\% & 23.8\% & 33.3\% & 33.3\%\\
In a hurry & 14.3\%  & 23.8\% & 38.1\% & 19\% & 4.8\%  \\
Long duration of red light & 14.3\%  & 14.3\% & 9.5\% & 33.3\% & 28.6\%  \\
Presence of other pedestrians who violate traffic signal & 9.5\%  & 9.5\% & 9.5\% & 19\% & 52.4\%  \\
Low traffic volume & 33.3\%  & 38.1\% & 14.3\% & 4.8\% & 9.5\%  \\
High traffic volume & 19\%  & 9.5\% & 0.0\% & 23.8\% & 47.6\%  \\
Police officer is on duty at the intersection & 23.8\%  & 4.8\% & 14.3\% & 9.5\% & 47.6\%  \\
 \hline
\end{tabular}
\caption{Probabilities of pedestrians' signal non-compliance under specific situations}
\label{table:AnswersProbabilities}
\end{table}

\begin{figure}[ht]
\centering
\includegraphics[width=0.8\textwidth]{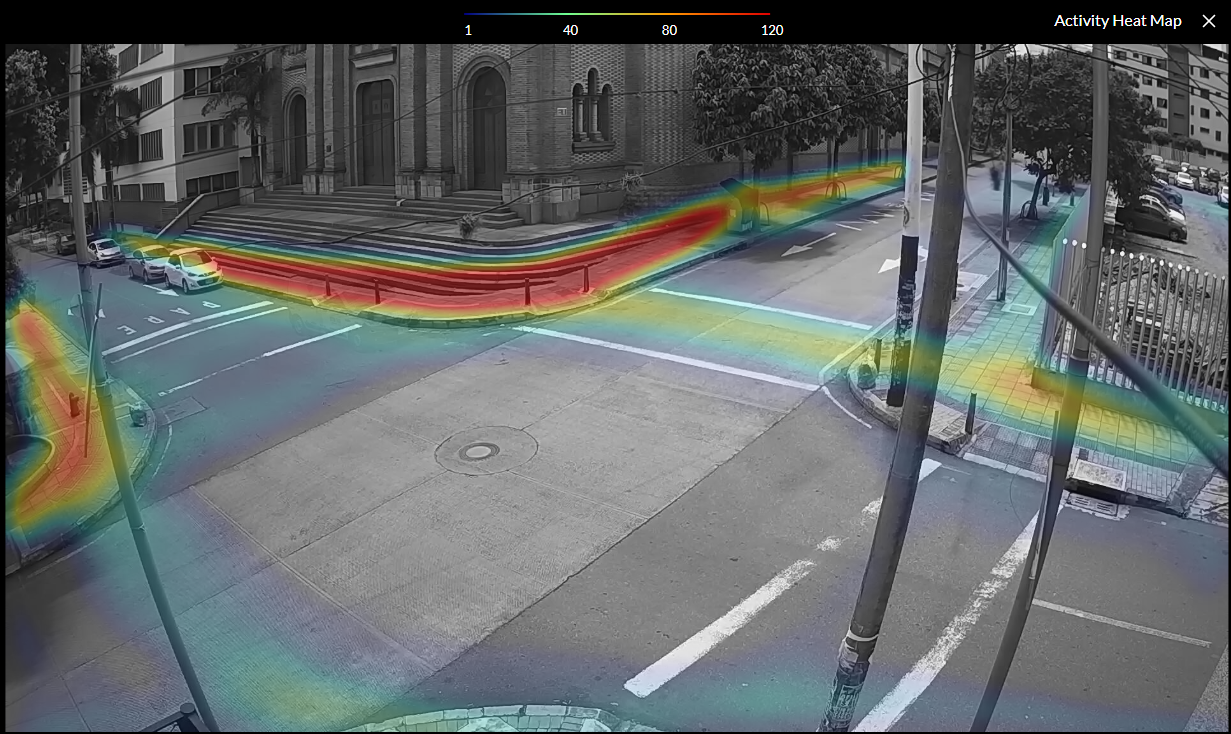}
\caption{Heat map showing where walkers go through}
\label{fig:i_camera_canaveral}
\end{figure}

The measured information for several video recordings, from the spot shown in Figure \ref{fig:i_moto_atravezada}, are presented in Table \ref{table:ListOfInformation}. The date and time were selected for working days, and the values shown in the table are the mean values during the week. 
The information was gathered from the videos thanks to the deep learning tool on Briefcam. We use this information as input for the simulations. 
Additionally, the same spot's heat-map is shown in Figure \ref{fig:i_camera_canaveral}. It reveals the frequent zones that pedestrians go, and that we use to narrow down the solutions to implement.

%%%%%% TABLE %%%%%%%
\begin{table}[h!]
\centering
\begin{tabular}{ | m{12em} | m{5em}| m{10em} | } 
 \hline
 \textbf{Variable} &  \textbf{Type} & \textbf{Observation}\\ 
 \hline\hline
 Date of Video  & 22/06/2019  &  \\
 Range hour &   e.g: 9:00 - 10:00 hrs & \\
 Pedestrians who cross the street  &   316  & 70\% man and 30\% woman \\
 Not safe events  &   0 & not registered in the video \\
 Pedestrians against the law  &   2 & \\
 Vehicles in the video  &  668 & \\ 
 Cars and motorbikes vehicles velocity average  &   60km/hr & \\

 \hline
\end{tabular}
\caption{The quantifiable information from the video recording}
\label{table:ListOfInformation}
\end{table}

With the additional information about the people's perception and the numeric data gathered from the videos, we have the micro-simulation input data. 
It is not possible to strictly imitate when walkers and vehicles appear, but they are defined by the number of objects that appear per minute. One frequent improvement in the pedestrians' care is the implementation of a bump. As a counterpart of that implementation, the queue of cars on the street produces traffic jams in the zone, altering the order and regular circulation. 
By changing the number of vehicles per minute, the simulation shows that maximum three vehicles stayed queuing in a row in an hour of simulation. One of the additional factors of accidents in a row is a pedestrian who walks seeing his cellphone, this increases the probability of accidents to 88\% for adults~\cite{Koopmans2015UrbanSeverity}. This number is possible to reproduce for all the pedestrians in the micro-simulation, then accidents appear. 

Overall, the road speed reducer decreases the speed to 2km/h from the 40km/h mandatory limit, and even from 60-70km/h in real-life events, according to the video records. From the results of the simulation, this small but significant change can save over 80\% of the people related in an accident - especially with the 88\% people presenting distraction or bravery conditions.

 \begin{figure}[ht]
\centering
\includegraphics[width=0.8\textwidth]{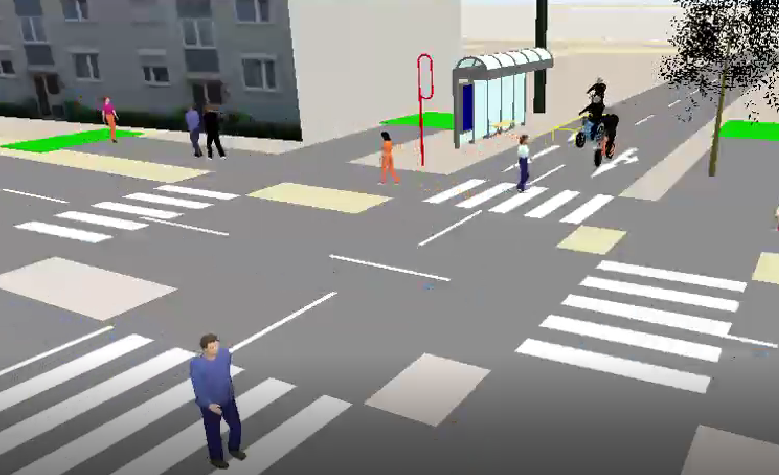}
\caption{Bump implemented in the simulation }
\label{fig:i_bump_implemented}
\end{figure}
 
%%%%%%%%%%%%%%%%%%%%%%%%%%%%%%%%%%%%%%%%%%%%%%%%%%%%%%%
\section{Conclusions}
The proposed model helps in determining the fittest variables that are more important in pedestrians' unsafe crossings in fast-growing cities. In other areas of a city or other cities having a similar environment, the model can be transferred to reduce the area's accidental rate.

Implementing a slight change in architecture - such as the speed bump deployment proposed in this article - produces a significant number of people's lives saved. 80\% of accident decrease can be achieved in our study.

\section{Future work}
The method proposed is a good first step to use infrastructure (cameras) and information (video-recordings) to build a smart city in the Bucaramanga case-study \cite{PuenM:TowSCICMFGUCTCS}. Our method analyzes different causes identified in the video recordings, other parameters such as the weather, building architectures, disabled people identification, vehicle conditions could be integrated and extend the method to global system dynamics. This pedestrian behaviour analysis can also provide urban services such as traffic optimization \cite{LebMA:ParLKGEUVT,LebMA:OnIRDMUVMT}, smart parking \cite{LinT:ASurSPS}, taxi recommendation \cite{QianS:SCRAMFTRR}, crisis management \cite{mouel2017decentralized}.

%%%%%%%%%%%%%%%%%%%%%%%%%%%%%%%%%%%%%%%%%%%%%%%%%%%%%%%
\section{Acknowledgements}
This work is supported by the Government and the Unidades Tecnologicas de Santander (project 879/2017). Thanks to the SC3-UIS Lab, the Colifri association, the CITI Lab at INSA Lyon and the CATAI workgroup - where this project was already discussed and received feedbacks.


\bibliographystyle{splncs04}
\bibliography{references2.bib}

\begin{thebibliography}{10}
\providecommand{\url}[1]{\texttt{#1}}
\providecommand{\urlprefix}{URL }
\providecommand{\doi}[1]{https://doi.org/#1}

\bibitem{Aaron2019}
Aar{\'{o}}n, M.A., G{\'{o}}mez, C.A., Fontalvo, J., G{\'{o}}mez, A.J.:
  {An{\'{a}}lisis de la Movilidad Vehicular en el Departamento de La Guajira
  usando Simulaci{\'{o}}n. El Caso de Riohacha y Maicao}. Informaci{\'{o}}n
  tecnol{\'{o}}gica  \textbf{30}(1),  321--332 (2019).
  \doi{10.4067/s0718-07642019000100321}

\bibitem{AlvarezPomar2016}
{\'{A}}lvarez-Pomar, L.: {Modelo de inteligencia colectiva de los sistemas
  peatonales}. Universidad Distrital Francisco Jose de Caldas  \textbf{23}(45),
   5--24 (2016)

\bibitem{Camara2018}
Camara, F., Giles, O., Madigan, R., Rothmuller, M., Rasmussen, P.H.,
  Vendelbo-Larsen, S.A., Markkula, G., Lee, Y.M., Garach, L., Merat, N., Fox,
  C.W.: {Predicting pedestrian road-crossing assertiveness for autonomous
  vehicle control}. IEEE Conference on Intelligent Transportation Systems,
  Proceedings, ITSC  \textbf{2018-Novem},  2098--2103 (2018).
  \doi{10.1109/ITSC.2018.8569282}

\bibitem{Campisi2018}
Campisi, T., Tesoriere, G., Canale, A.: {The pedestrian micro-simulation
  applied to the river Neretva: The case study of the Mostar "old bridge"}. AIP
  Conference Proceedings  \textbf{2040}(November) (2018).
  \doi{10.1063/1.5079193}

\bibitem{Cantillo2015}
Cantillo, V., Arellana, J., Rolong, M.: {Modelling pedestrian crossing
  behaviour in urban roads: A latent variable approach}. Transportation
  Research Part F: Traffic Psychology and Behaviour  \textbf{32},  56--67 (7
  2015). \doi{10.1016/j.trf.2015.04.008},
  \url{https://linkinghub.elsevier.com/retrieve/pii/S1369847815000716}

\bibitem{Chen2016}
Chai, C., Shi, X., Wong, Y.D., Er, M.J., Gwee, E.T.M.: {Fuzzy logic-based
  observation and evaluation of pedestrians' behavioral patterns by age and
  gender}. Transportation Research Part F: Traffic Psychology and Behaviour
  \textbf{40},  104--118 (2016). \doi{10.1016/j.trf.2016.04.004},
  \url{http://dx.doi.org/10.1016/j.trf.2016.04.004}

\bibitem{Cookson2011}
Cookson, R., Richards, D., Cuerden, R.: {The characteristics of pedestrian road
  traffic accidents and the resulting injuries} (2011)

\bibitem{das2005}
Das, S., Manski, C.F., Manuszak, M.D.: {Walk or wait? An empirical analysis of
  street crossing decisions}. Journal of Applied Econometrics  \textbf{20}(4),
  529--548 (2005). \doi{10.1002/jae.791}

\bibitem{Helbing1995}
Helbing, D., Moln{\'{a}}r, P.: {Social force model for pedestrian dynamics}.
  Physical Review E  \textbf{51}(5),  4282--4286 (1995).
  \doi{10.1103/PhysRevE.51.4282}

\bibitem{Holland2010}
Holland, C., Hill, R.: {Gender differences in factors predicting unsafe
  crossing decisions in adult pedestrians across the lifespan: A simulation
  study}. Accident Analysis and Prevention  \textbf{42}(4),  1097--1106 (2010).
  \doi{10.1016/j.aap.2009.12.023},
  \url{http://dx.doi.org/10.1016/j.aap.2009.12.023}

\bibitem{Koopmans2015UrbanSeverity}
Koopmans, J.M., Friedman, L., Kwon, S., Sheehan, K.: {Urban crash-related child
  pedestrian injury incidence and characteristics associated with injury
  severity}. Accident Analysis and Prevention  \textbf{77},  127--136 (4 2015).
  \doi{10.1016/j.aap.2015.02.005}

\bibitem{Kouabenan2005}
Kouabenan, D.R., Guyot, J.M.: {Study of the causes of pedestrian accidents by
  severity}. Journal of Psychology in Africa  \textbf{14}(2) (2005).
  \doi{10.4314/jpa.v14i2.30620}

\bibitem{Kumar2011}
Kumar, N.., {G.Umadevi}: {Application of System Dynamic Simulation Modeling in
  Road Safety}. Trasnportation Research Board  (2011)

\bibitem{mouel2017decentralized}
{Le Mouël}, F., Hernández, C.B., Carrillo, O., Pedraza, G.: Decentralized,
  robust and efficient services for an autonomous and real-time urban crisis
  management (2017), \url{https://arxiv.org/abs/1703.04519}

\bibitem{LebMA:OnIRDMUVMT}
L{\`e}bre, M.A., {Le Mou{\"e}l}, F., M{\'e}nard, E.: On the importance of real
  data for microscopic urban vehicular mobility trace. In: Proceedings of the
  14th International Conference on ITS Telecommunications (ITST'2015). pp.
  22--26. Copenhagen, Denmark (Dec 2015). \doi{10.1109/ITST.2015.7377394}

\bibitem{LebMA:ParLKGEUVT}
L{\`e}bre, M.A., {Le Mou{\"e}l}, F., M{\'e}nard, E.: Partial and local
  knowledge for global efficiency of urban vehicular traffic. In: Proceedings
  of the IEEE 82nd Vehicular Technology Conference (VTC'2015-Fall). pp.~1--5.
  Boston, MA, USA (Sep 2015). \doi{10.1109/VTCFall.2015.7391065}

\bibitem{LinT:ASurSPS}
Lin, T., Rivano, H., {Le Mou{\"e}l}, F.: A survey of smart parking solutions.
  IEEE Transactions on Intelligent Transportation Systems  \textbf{18}(12),
  3229--3253 (2017). \doi{10.1109/TITS.2017.2685143}

\bibitem{Mendez2014}
M{\'{e}}ndez-Giraldo, G., {\'{A}}lvarez-Pomar, L.: {Dynamic model to analyze
  pedestrian traffic policies in Bogota}. Dyna  \textbf{81}(186), ~276 (2014).
  \doi{10.15446/dyna.v81n186.45219}

\bibitem{oskarbski2018}
Oskarbski, J., Gumi{\'{n}}ska, L.: {The application of microscopic models in
  the study of pedestrian traffic}. MATEC Web of Conferences  \textbf{231},
  ~1--7 (2018). \doi{10.1051/matecconf/201823103003}

\bibitem{Pau2018}
Pau, G., Campisi, T., Canale, A., Severino, A., Collotta, M., Tesoriere, G.:
  {Smart pedestrian crossing management at traffic light junctions through a
  fuzzy-based approach}. Future Internet  \textbf{10}(2) (2018).
  \doi{10.3390/fi10020015}

\bibitem{PuenM:TowSCICMFGUCTCS}
Puentes, M., Arroyo, I., Carrillo, O., Barrios, C.J., {Le Mou{\"e}l}, F.:
  Towards smart-city implementation for crisis management in fast-growing and
  unplanned cities: The colombian scenario. {Ingenier{\'i}a y Ciencia}
  \textbf{16}(32),  151--169 (Nov 2020). \doi{10.17230/ingciencia.16.32.7}

\bibitem{PuenM:PedBMSRTDI}
Puentes, M., Novoa, D., Nivia, J.D., Hernndez, C.B., Carrillo, O., {Le
  Mou{\"e}l}, F.: Pedestrian behaviour modeling and simulation from real time
  data information. In: 2nd Workshop CATA{\"I} - SmartData for Citizen
  Wellness. Bogot{\'a}, Colombia (Oct 2019),
  \url{http://www.catai.fr/catai2019/}

\bibitem{QianS:SCRAMFTRR}
Qian, S., Cao, J., {Le Mou{\"e}l}, F., Sahel, I., Li, M.: {SCRAM}: A sharing
  considered route assignment mechanism for fair taxi route recommendations.
  In: Proceedings of the 21st ACM SIGKDD Conference on Knowledge Discovery and
  Data Mining (KDD'2015). pp. 955--964. Sydney, Australia (Aug 2015).
  \doi{10.1145/2783258.2783261}

\bibitem{Rasouli2018}
Rasouli, A., Kotseruba, I., Tsotsos, J.K.: {Understanding Pedestrian Behavior
  in Complex Traffic Scenes}. IEEE Transactions on Intelligent Vehicles
  \textbf{3}(1),  61--70 (3 2018). \doi{10.1109/tiv.2017.2788193},
  \url{http://ieeexplore.ieee.org/document/8241847/}

\bibitem{teknomo2016}
Teknomo, K., Takeyama, Y., Inamura, H.: {Review on Microscopic Pedestrian
  Simulation Model} (March), ~1--2 (2016),
  \url{http://arxiv.org/abs/1609.01808}

\bibitem{TorG:CiuRSCPDAVCP}
Torres, G., al.: La ciudad -- regi{\'o}n sostenible como proyecto: desaf\'{i}os
  actuales. Visiones cruzadas y perspectivas. Universidad Nacional de Colombia,
  Sede Bogot\'{a}, Bogot\'{a} (Oct 2019),
  \url{http://bdigital.unal.edu.co/74466/}

\bibitem{Ulfarsson2010}
Ulfarsson, G.F., Kim, S., Booth, K.M.: {Analyzing fault in pedestrian-motor
  vehicle crashes in North Carolina}. Accident Analysis and Prevention
  \textbf{42}(6),  1805--1813 (2010). \doi{10.1016/j.aap.2010.05.001},
  \url{http://dx.doi.org/10.1016/j.aap.2010.05.001}

\bibitem{Yang2006q}
Yang, J., Deng, W., Wang, J., Li, Q., Wang, Z.: {Modeling pedestrians' road
  crossing behavior in traffic system micro-simulation in China}.
  Transportation Research Part A: Policy and Practice  \textbf{40}(3),
  280--290 (2006). \doi{10.1016/j.tra.2005.08.001}

\bibitem{Zhang2014}
Zhang, G., Yau, K.K., Zhang, X.: {Analyzing fault and severity in
  pedestrian-motor vehicle accidents in China}. Accident Analysis and
  Prevention  \textbf{73},  141--150 (2014). \doi{10.1016/j.aap.2014.08.018},
  \url{http://dx.doi.org/10.1016/j.aap.2014.08.018}

\end{thebibliography}

\end{document}